\documentclass[sigconf]{aamas} 

\setcopyright{ifaamas}
\acmConference[AAMAS '21]{Proc.\@ of the 20th International Conference on Autonomous Agents and Multiagent Systems (AAMAS 2021)}{May 3--7, 2021}{Online}{U.~Endriss, A.~Now\'{e}, F.~Dignum, A.~Lomuscio (eds.)}
\copyrightyear{2021}
\acmYear{2021}
\acmDOI{}
\acmPrice{}
\acmISBN{}

\usepackage{booktabs}
\usepackage[utf8]{inputenc} 
\usepackage{amsmath} 

\usepackage{algorithmic}
\usepackage{algorithm}

\newcommand{\startpara}[1]{{\vskip1pt\noindent{\bf #1.}}}

\acmSubmissionID{280}

\title{Safe Multi-Agent Reinforcement Learning via Shielding} 

\author{Ingy ElSayed-Aly}
\affiliation{
\institution{University of Virginia}
\city{Charlottesville}
\state{Virginia}
\postcode{22904}
}
\email{ie3ne@virginia.edu}

\author{Suda Bharadwaj}
\affiliation{
\institution{University of Texas at Austin}
\city{Austin}
\state{Texas}
\postcode{78712-1221}
}
\email{suda.b@utexas.edu}

\author{Christopher Amato}
\affiliation{
\institution{Northeastern University}
\city{Boston}
\state{Massachusetts}
\postcode{02115}
}
\email{c.amato@northeastern.edu}

\author{Rüdiger Ehlers}
\affiliation{
\institution{Clausthal University of Technology}
\city{Clausthal-Zellerfeld}
\country{Germany}
\postcode{38678}
}
\email{ruediger.ehlers@tu-clausthal.de}

\author{Ufuk Topcu}
\affiliation{
\institution{University of Texas at Austin}
\city{Austin}
\state{Texas}
\postcode{78712-1221}
}
\email{utopcu@utexas.edu}

\author{Lu Feng}
\affiliation{
\institution{University of Virginia}
\city{Charlottesville}
\state{Virginia}
\postcode{22904}
}
\email{lu.feng@virginia.edu}

\begin{abstract}

Multi-agent reinforcement learning (MARL) has been increasingly used in a wide range of safety-critical applications, which require  guaranteed safety (e.g., no unsafe states are ever visited) during the learning process.
Unfortunately, current MARL methods do not have safety guarantees. 
Therefore, we present two shielding approaches for safe MARL. 
In \emph{centralized shielding}, we synthesize a single shield to monitor all agents' joint actions and correct any unsafe action if necessary.
In \emph{factored shielding}, we synthesize multiple shields based on a factorization of the joint state space observed by all agents; the set of shields monitors agents concurrently and each shield is only responsible for a subset of agents at each step.
Experimental results show that both approaches can guarantee the safety of agents during learning without compromising the quality of learned policies; moreover, factored shielding is more scalable in the number of agents than centralized shielding.
\end{abstract}
    
      
\begin{document}

    
    \pagestyle{fancy}
    \fancyhead{}
	
	\maketitle

	
	\section{Introduction}

Multi-agent reinforcement learning (MARL) addresses sequential decision-making problems where multiple agents interact with each other in a common environment. In recent years, MARL methods have been increasingly used in a wide range of safety-critical applications from traffic management~\cite{singh2020hierarchical} to robotic control~\cite{yu2019coordinated} to autonomous driving~\cite{shalev2016safe}.
Existing MARL methods~\cite{hernandez2019survey,zhang2019multi} focus mostly on optimizing policies based on returns, none of which can guarantee safety (e.g., no unsafe states are ever visited) during the learning process. 
Nevertheless, learning with provable safety guarantees is necessary for many safety-critical MARL applications where the agents (e.g., robots, autonomous cars) may break during the exploration process and lead to catastrophic outcomes. 

A recent work~\cite{alshiekh2018safe} developed a shielding framework for single-agent reinforcement learning (RL), which synthesizes a shield to enforce the correctness of safety specifications in linear temporal logic (LTL)~\cite{pnueli1977temporal}.  
The shield guarantees safety during learning by monitoring the RL agent's actions and preventing the exploration of any unsafe action that violates the LTL safety specification. 
In this paper, we adapt the shielding framework to the multi-agent setting. Guaranteeing safety for multiple agents with potentially competing goals is more challenging than the single-agent setting, because safety is an emergent property that concerns the coupling of all agents. In addition, the combinatorial nature of MARL (i.e., the joint state space and joint action space increase exponentially with the number of agents) poses scalability issues to the computation of shields. 

We present in this paper the first work to provide safety guarantees (expressed as LTL specifications) for MARL.
Our contributions are threefold. First, we develop a \emph{centralized shielding} approach for MARL, where we synthesize a single shield to centrally monitor the joint actions of all agents. The shield determines that a joint action is safe if all agents satisfy the safety specification. We follow the \emph{minimal interference} principle proposed in \cite{alshiekh2018safe}; that is, a shield should restrict the agents as infrequently as possible and only corrects the actions that violate the safety specification. Moreover, we introduce an additional interpretation of minimal interference in the multi-agent setting: a shield should change the actions of as few agents as possible when correcting an unsafe joint action. 
The centralized shielding approach has limited scalability, because the computational cost of synthesizing shields depends on the number of MARL agents and the complexity of the safety specification.

Second, we develop a \emph{factored shielding} approach for MARL to address the aforementioned scalability issues. 
The factored shielding offers a divide-and-conquer approach: multiple shields are computed based on a factorization of the joint state space observed by all agents. The set of factored shields monitors agents concurrently and each shield is only responsible for a subset of agents at each step.  
Agents can join or leave a factored shield at any time depending on their states.
Factored shields enforce the correctness of safety specification by preventing unsafe actions similarly to the centralized shield.
While each individual factored shield can only monitor a limited number of agents due to the restriction of shield computation, we can employ as many shields as needed; and together the set of factored shields can monitor a large number of MARL agents.

Third, we showcase the performance of the two shielding approaches via experimental evaluation on six benchmark problems in a grid world~\cite{melo2009learning} and a cooperative navigation~\cite{yang2019cm3} environment. 
We used two MARL algorithms, CQ-learning~\cite{de2010learning} and MADDPG~\cite{lowe2017multi}, in our experiments to demonstrate that the shielding approaches are compatible with different MARL algorithms.
Experimental results show that the two shielding approaches can both guarantee the safety of agents during learning without compromising the quality of learned policies; moreover, factored shielding is more scalable in the number of agents than centralized shielding.

	\section{Related Work} 
Safe reinforcement learning (RL) is an active research area, but existing results focus mostly on the single-agent setting~\cite{garcia2015comprehensive}, while safe MARL is still a relatively uncharted territory~\cite{zhang2019multi}. 
To the best of our knowledge, this paper presents the first safety-constrained MARL method.
The survey in~\cite{garcia2015comprehensive} classifies safe RL methods into two categories: (1) transforming the optimization criterion with a safety factor, such as the worst case criterion, risk-sensitive criterion, or constrained criterion; and (2) modifying the exploration process through the incorporation of external knowledge (e.g., demonstrations, teacher advice) or the guidance of a risk metric. 
Our shielding approaches fall into the second category.
In particular, shields act similarly to a teacher who provides information (e.g., safe actions) to the learner when necessary (e.g., unsafe situations are detected).
The concept of shielding was introduced to RL for the single-agent setting in~\cite{alshiekh2018safe}.
In this work, we adapt the shielding framework for MARL via addressing challenges such as the coupling of agents and scalabilty issues in the multi-agent setting.

Different safety objectives for RL have been considered in the literature, such as the variance of the return, or limited visits of error states~\cite{garcia2015comprehensive}. In this work, we synthesize shields that enforce safety specifications expressed in linear temporal logic (LTL)~\cite{pnueli1977temporal}, which is a commonly used specification language in formal methods for safety-critical systems~\cite{alur2015principles,baier2008principles}.
For example, LTL has been used to express complex task specifications for robotic planning and control~\cite{kress2009temporal,ulusoy2013optimality}. 
Several recent works~\cite{hasanbeig2020cautious,bozkurt2020control,hahn2019omega} have developed reward shaping techniques that translate logical constraints expressed in LTL to reward functions for RL.
However, as we demonstrated in our experiments (Section~\ref{sec:exp}), relying on reward functions only is not sufficient for MARL methods to learn policies that guarantee the safety (e.g., no collisions).


The shield synthesis technique based on solving two-player safety games was developed in~\cite{bloem2015shield} for enforcing safety properties of a system at runtime, and was adopted in~\cite{alshiekh2018safe} to synthesize shields for single-agent RL.
We further adapt this technique to synthesize centralized and factored shields for MARL in this paper. 
There are a few recent works~\cite{raju2019decentralized, bharadwaj2019synthesis} considering the shield synthesis for multi-agent (offline) planning and coordination, none of which are directly applicable for MARL. 


	
	\section{Background} \label{sec:background}
	
A discrete probability \emph{distribution} over a (countable) set $S$ is
a function $\mu : S \to [0, 1]$ such that $\sum_{s \in S} \mu(s) =  1$.
Let $Distr(S)$ denote the set of distributions over $S$.
We use $\mathbb{R}$ to denote the real numbers. 
Given an alphabet $\Sigma$, we denote by $\Sigma^\omega$ and $\Sigma^*$ the set of infinite and finite words over $\Sigma$, respectively. 

\startpara{Multi-Agent Reinforcement Learning (MARL)}
We follow the Markov game formulation of MARL in~\cite{zhang2019multi}.
A \emph{Markov game} is a tuple $(N, S, \{A^i\}_{i \in N}, P, \{R^i\}_{i \in N}, \gamma)$
with a finite set $N=\{1,\cdots,n\}$ of agents, and a finite state space $S$ observed by all agents;
let $A:=A^1 \times \cdots \times A^n$ be the set of joint actions for all agents, where $A^i$ denotes the actions of agent $i \in N$;
the probabilistic transition function $P: S \times A \to Distr(S)$ is defined over the joint states and actions of all agents;
$R^i: S \times A \times S \to \mathbb{R}$ is an immediate reward function for agent $i$ under the joint states and actions;
$\gamma \in [0,1]$ is the discount factor of future rewards. 
At time step $t$, each agent chooses an action $a^i_t \in A^i$ based on the observed state $s_t \in S$. 
The environment moves to state $s_{t+1}$ with the probability $P(s_t, a_t, s_{t+1})$, 
where $a_t=(a^1_t,\cdots,a^n_t)$ is the joint action of all agents,
and rewards agent $i$ with $R^i(s_t, a_t, s_{t+1})$.
The goal of an individual agent $i$ is to learn a policy $\pi^i: S \to Distr(A^i)$ that optimizes the expectation of cumulative future rewards $\mathbb{E}[\sum_{t=0}^{\infty}\gamma^t R^i(s_t, a_t, s_{t+1})]$.
The performance of individual agent $i$ is not only influenced by its own policy, but also the choices of all other agents.

Depending on agents' goals, MARL algorithms can be categorized as fully cooperative (i.e., agents collaborate to optimize a common long-term return), fully competitive (i.e., zero-sum game among agents), or a mixed setting that involves both cooperative and competitive agents.
In our experiments (Section~\ref{sec:exp}), we used the following three mixed-setting algorithms. 
Independent Q-learning~\cite{tan1993multi} is a baseline algorithm where agents learn Q-values over their own action set independently and do not use any information about other agents.
CQ-learning~\cite{de2010learning} is an algorithm that allows agents to act independently most of the time and only accounts for the other agents when necessary (e.g., when conflict situations are detected).
MADDPG~\cite{lowe2017multi} is a deep MARL algorithm featuring centralized training with decentralized execution, in which each agent trains models simulating each of the other agents' policies based on its observation of their actions.

Scalability is a key challenge of MARL, due to its combinatorial nature. 
For example, our experiments can only use two agents with CQ-learning, but more than four agents with MADDPG which applies deep neural networks for function approximation to mitigate the scalability issue. 
Another key challenge of MARL is the lack of convergence guarantees in general, except for some special settings~\cite{zhang2019multi}. 
As multiple agents learn and act concurrently, the environment faced by an individual agent becomes non-stationary, which invalidates the stationary assumption used for proving convergence in single-agent RL algorithms.

\startpara{Safety Specifications and Safety Games}
We use linear temporal logic (LTL)~\cite{pnueli1977temporal} to express safety specifications. 
In addition to propositional logical operators, LTL employs temporal operators such as
$\bigcirc$ (next), $\mathsf{U}$ (until), $\Box$ (always), and $\Diamond$ (eventually).
The set of words that satisfies an LTL formula $\phi$ represents a language $\mathcal{L}(\phi) \subseteq (2^{\mathsf{AP}})^\omega$, where $\mathsf{AP}$ is a given set of atomic propositions.
LTL formulas can be used to express a wide variety of requirements. 
We focus on safety specifications, which are informally interpreted as ``something bad should never happen''. 
For example, the LTL formula $\Box \neg \mathsf{unsafe}$ expresses that ``unsafe states should never be visited''. An LTL safe specification can be translated into a safe language accepted by a deterministic finite automaton (DFA)~\cite{kupferman2001model}.

Formally, a \emph{deterministic finite automaton} is a tuple $(Q, q_0, \Sigma, \delta, F)$ with a finite set of states $Q$, an initial state $q_0 \in Q$, a finite alphabet $\Sigma$, the transition function $\delta: Q \times \Sigma \to Q$, and a finite set of accepting states $F \subseteq Q$. 
Let $q_0\sigma_0q_1\sigma_1 \dots \in (Q \times \Sigma)^\omega$ be a run of the DFA.
The word $\sigma_0\sigma_1 \dots$ is in the safety language accepted by the DFA if the run only visits accepting states of the DFA, i.e., $q_i \in F$ for all $i \ge 0$.

We use Mealy machines to represent shields. Formally, a \emph{Mealy machine} is a tuple $(Q, q_0, \Sigma_I, \Sigma_O, \delta, \lambda)$ with a finite set of states $Q$, an initial state $q_0 \in Q$, finite sets of input alphabet $\Sigma_I$ and output alphabet $\Sigma_O$, the transition function $\delta: Q \times \Sigma_I \to Q$, and the output function $\lambda: Q \times \Sigma_I \to \Sigma_O$.
For a given input trace $\sigma_0\sigma_1 \dots \in \Sigma^{\omega}_I$, the Mealy machine generates a corresponding output trace $\lambda(q_0,\sigma_0)\lambda(q_1,\sigma_1)\dots \in \Sigma^{\omega}_O$ where
$q_{i+1}=\delta(q_i,\sigma_i)$ for all $i\ge 0$.

As we will describe later, we synthesize shields by solving two-player safety games. 
Formally, a \emph{two-player safety game} is a tuple $(G, g_0, \Sigma_1, \Sigma_2, \delta, F)$ with a finite set of game states $G$, an initial state $g_0 \in G$, finite sets of alphabet $\Sigma_1$ and $\Sigma_2$ for Player 1 and Player 2 respectively, the transition function $\delta: G \times \Sigma_1 \times \Sigma_2 \to G$, and a set of safe states $F \subseteq G$ defines the \emph{winning condition} such that a play $g_0g_1\dots$ of the game is winning iff $g_i \in F$ for all $i \ge 0$.
At each game state $g_i \in G$, Player 1 chooses an action $a_i^1 \in \Sigma_1$, 
then Player 2 chooses an action $a_i^2 \in \Sigma_2$, and the game moves to the next state $g_{i+1}=\delta(g_i,a_i^1,a_i^2)$.
A memoryless strategy for Player 2 is a function $\kappa: G \times \Sigma_1 \to \Sigma_2$.
A \emph{winning region} $W \subseteq F$ is the set of states from which there exists a winning strategy (i.e., all plays constructed using the strategy satisfy the winning condition).

	\section{Centralized Shielding} \label{sec:centralized}
	
We introduce a \emph{centralized shield} (i.e., a single shield for all agents) into the traditional MARL process. In the following, we first describe how the centralized shield interacts with the learning agents and the environment to achieve safe MARL, then we present our method for synthesizing the centralized shield. 

\begin{figure}[t]
	\centering
	\includegraphics[width=0.7\columnwidth]{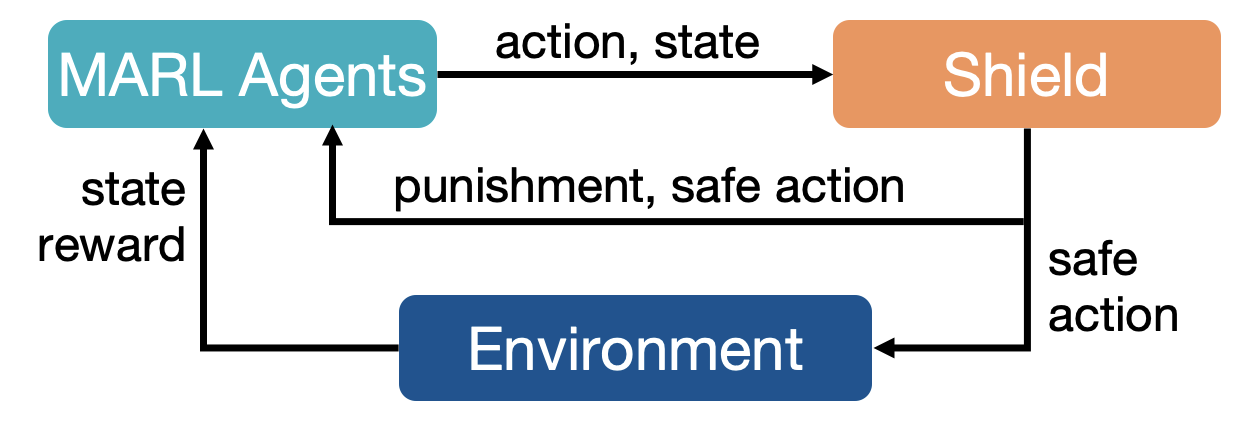}
	\caption{Safe MARL with centralized shielding.}
	\label{fig:centralized}
\end{figure}

Figure~\ref{fig:centralized} illustrates the interaction of the centralized shield, the MARL agents, and the environment. 
Algorithm~\ref{code_centralized} summarizes the centralized shield's behavior at time step $t$. 
The shield monitors the joint action $a_t=(a^1_t,\cdots,a^n_t)$ chosen by the MARL agents.
If the shield detects that $a_t$ is unsafe (i.e., violates the safety specification) at the agents' joint state $s_t \in S$, the shield substitutes $a_t$ with a safe joint action $\bar{a}_t$; 
otherwise, the shield forwards $a_t$ to the environment directly (i.e., $\bar{a}_t = a_t$).
The environment receives the action $\bar{a}_t$ output by the shield, moves to state $s_{t+1} \in S$, and provides reward $R^k(s_t, \bar{a}_t, s_{t+1})$ for each agent $k$ to update its policy. 
Meanwhile, the shield assigns a punishment $\rho^k_t$ to agent $k$ (where $\bar{a}^k_t \neq a^k_t$) to help the MARL algorithm learn about the cost of unsafe actions.

%
A centralized shield enforces the safety specification during the learning process (i.e., any unsafe action is corrected to a safe action before being sent to the environment). 
Moreover, we require the shield to restrict MARL agents as rarely as possible via the  \emph{minimal interference} criteria:
(1) the shield only corrects the joint action $a_t$ if it violates the safety specification,
and (2) the shield seeks a safe joint action $\bar{a}_t$ that changes as few of the agents' actions as possible from $a_t$.

Our approach synthesizes a centralized shield based on the safety specification and a coarse environment abstraction. 
Note that we do not require the environment dynamics to be completely known in advance. 
The shield can be synthesized based on a coarse abstraction of the environment that is sufficient to reason about the potential violations of safety specifications. 
For example, before deploying a team of robots for a disaster search and rescue mission, we may use some low-resolution satellite imagery to build a coarse, high-level abstraction about the terrain environment for shield synthesis. However, such a coarse environment abstraction is not sufficient for planning algorithms that rely on complete models of the environment. Therefore, MARL agents still need to learn about the concrete environment dynamics. 

\begin{algorithm}[t]
\caption{Centralized shielding at time step $t$} \label{code_centralized}
\algsetup{linenosize=\tiny}
\small
\begin{algorithmic} [1]
\REQUIRE Shield $\mathcal{S}$, MARL agents' joint action $a_t=(a^1_t,\cdots,a^n_t)$ and joint state $s_t=(s^1_t,\cdots,s^n_t)$, a constant punishment cost $c$
\ENSURE Safe joint action $\bar{a}_t$, punishment $\rho_t$
\STATE $\rho_t \leftarrow 0$
\STATE $\bar{a} \leftarrow$ safe action output by the shield $\mathcal{S}$
\FORALL{agent $k$ such that $\bar{a}^k \neq a^k$}
    
    \STATE $\rho^k_t \leftarrow c$
\ENDFOR

\RETURN $\bar{a}_t$, $\rho_t$
\end{algorithmic}
\end{algorithm}

We describe how to synthesize centralized shields as follows. 
We assume some coarse environment abstraction has been given as a DFA $\mathcal{A}^e=(Q^e, q^e_0, \Sigma^e, \delta^e, F^e)$ with the alphabet $\Sigma^e = L \times A$, where an observation function $f: S \to L$ maps the MARL agents' joint state space $S$ to some observation set $L$, and $A$ is the joint action set of all agents. 
We translate the safety specification expressed as an LTL formula to another DFA $\mathcal{A}^s=(Q^s, q^s_0, \Sigma^s, \delta^s, F^s)$ with the same alphabet $\Sigma^s = L \times A$.
We combine $\mathcal{A}^e$ and $\mathcal{A}^s$ into a two-player safety game 
$\mathcal{G}=(G, g_0, \Sigma_1, \Sigma_2, \delta^g, F)$
where $G=Q^e \times Q^s$, $g_0 = (q^e_0,q^s_0)$, $\Sigma_1 = L$, $\Sigma_2 = A$, 
$\delta^g((q^e, q^s), l, a) = (\delta^e(q^e, (l \times a)), \delta^s(q^s, (l \times a))$ for all $(q^e, q^s) \in G$, $l \in L$, and $a \in A$, and $F = Q^e \times F^s$.
We solve the two-player safety game $\mathcal{G}$ and compute the wining region $W \subseteq F$ using the techniques described in~\cite{bloem2015shield}.
We construct the centralized shield represented as a Mealy machine $\mathcal{S}=(Q, q_0, \Sigma_I, \Sigma_O, \delta, \lambda)$,
where the state space is given by the game states $Q=G=Q^e \times Q^s$,
the initial state $q_0 = g_0 = (q^e_0,q^s_0)$,
the input alphabet $\Sigma_I = L \times A$,
the output alphabet $\Sigma_O = A$;
the transition function $\delta(g,(l,a))=\delta^g(g, l, \lambda(g,(l,a)))$ for all $g \in G$, $l \in L$, and $a \in A$;
the output function $\lambda(g,(l,a))=a$ if $\delta^g(g, l, a) \in W$, 
and $\lambda(g,(l,a))=\bar{a}$ if $\delta^g(g, l, a) \not\in W$, where $\bar{a} \in A$ is a safe action with $\delta^g(g, l, \bar{a}) \in W$ and only differs from the unsafe action $a$ in terms of the minimal number of agents' actions. 
We also define a (negative) constant $c$ as punishment for unsafe actions.  
The computational cost of synthesizing centralized shields grows exponentially as the number of agents increases, and also depends on the complexity of the safety specification and environment abstraction.

\begin{figure}[t]
	\centering
	\includegraphics[width=0.7\columnwidth]{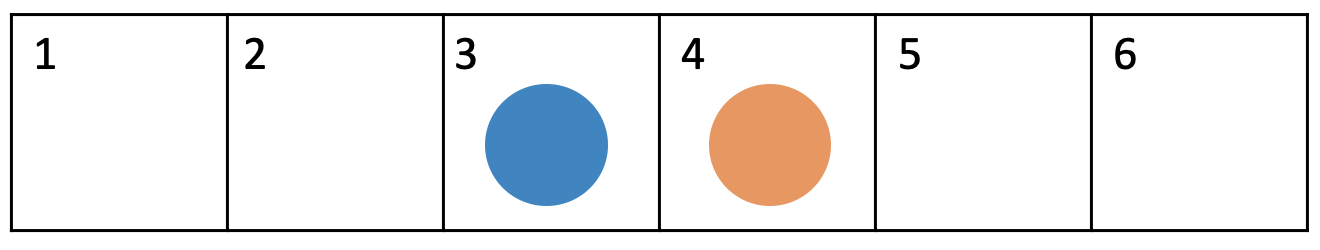}
	\caption{Example grid map with two agents. }
	\label{fig:grid}
\end{figure}

\begin{figure}[t]
	\centering
	\includegraphics[width=\columnwidth]{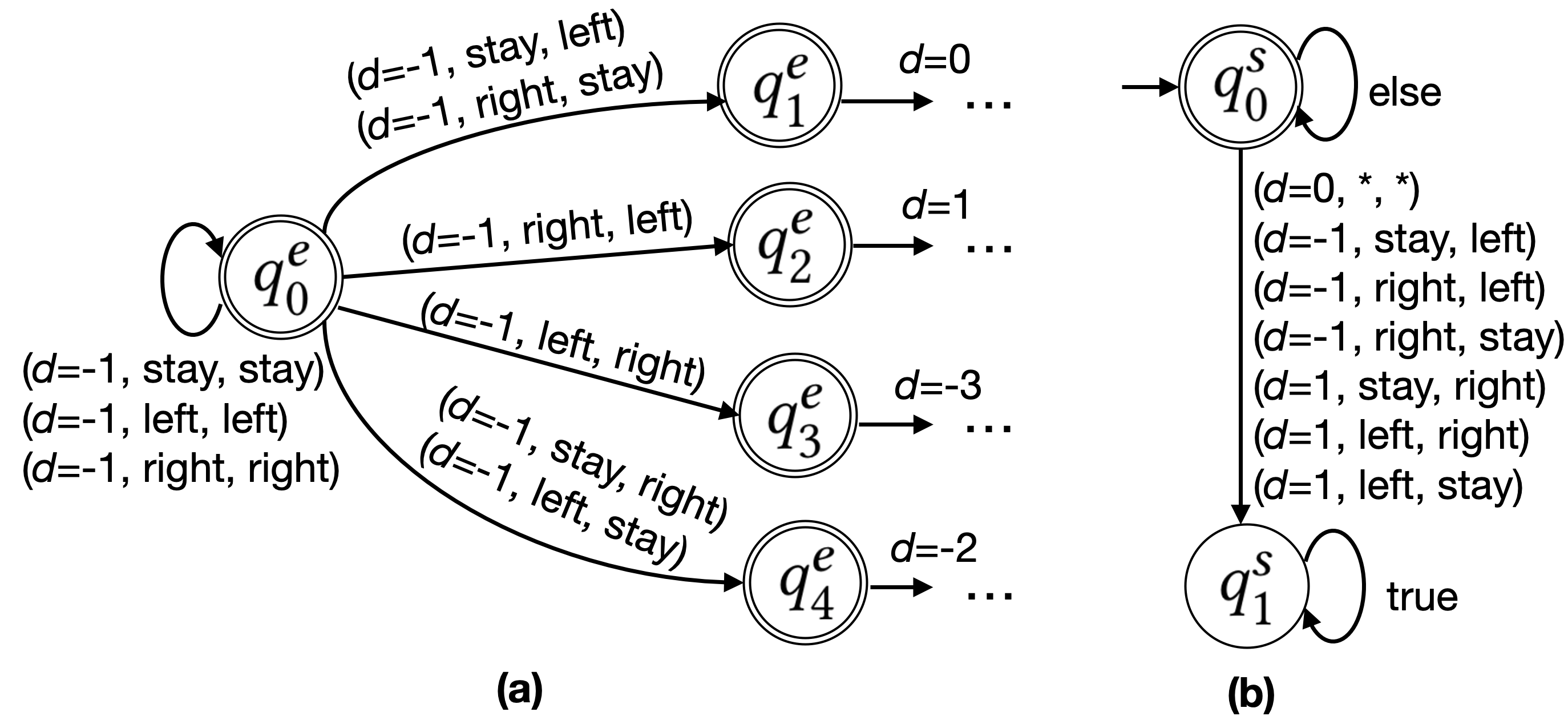}
	\caption{(a) An example environment abstraction DFA $\mathcal{A}^e$. (b) An example safety specification DFA $\mathcal{A}^s$. (Double circle denotes accepting states of DFAs. * refers to any action.)}
	\label{fig:eg_centralized}
\end{figure}

\begin{figure}[t]
	\centering
	\includegraphics[width=0.7\columnwidth]{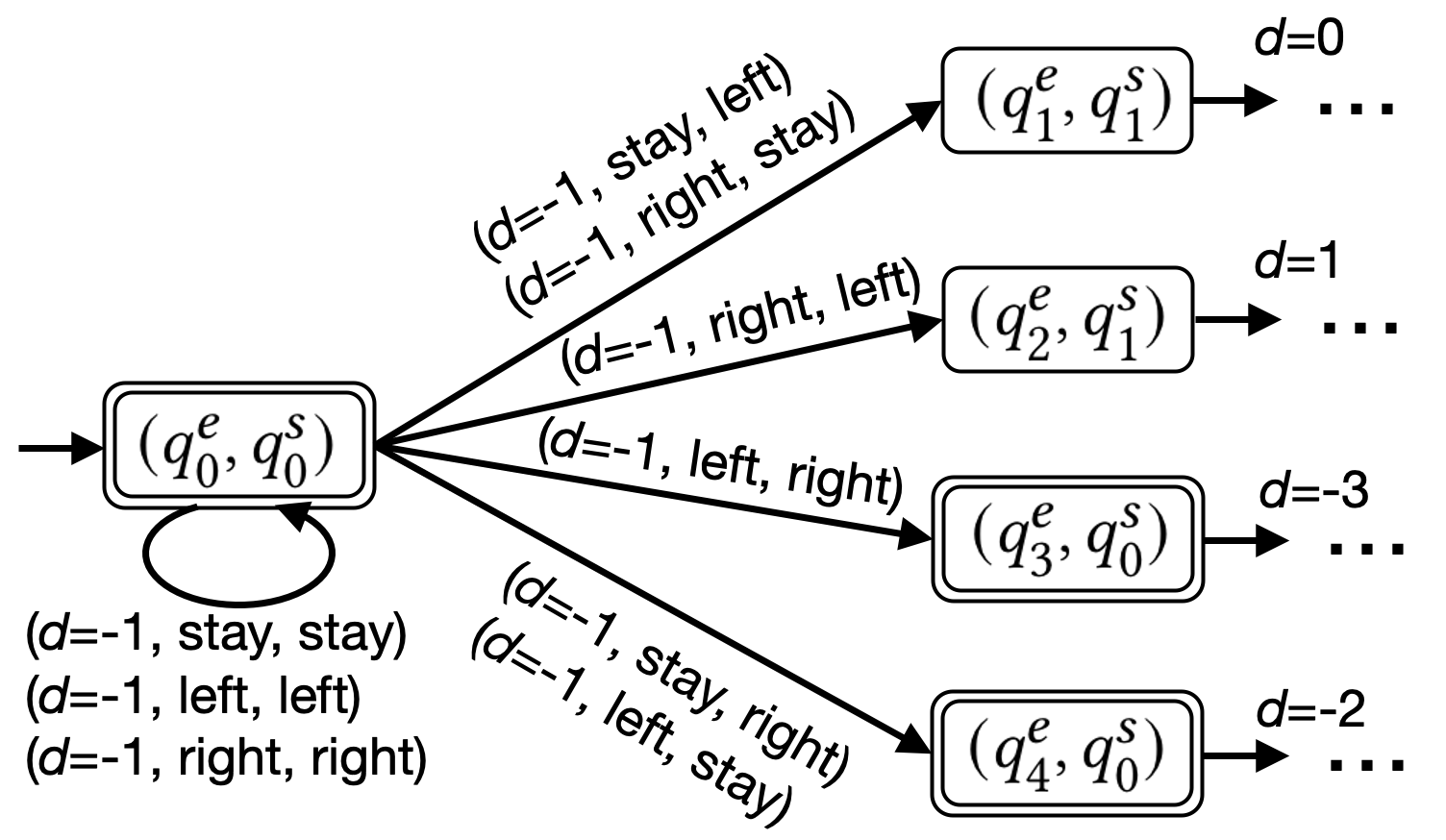}
	\caption{An example safety game given by the product of $\mathcal{A}^e$ and $\mathcal{A}^s$ shown in Figure~\ref{fig:eg_centralized}. Double circles denote safe states. }
	\label{fig:eg_game}
\end{figure}

To exemplify the shield synthesis method, let us consider two agents (blue and orange) in the grid map shown in Figure~\ref{fig:grid}. Each agent can move left or right, or stay in the same grid. An agent receives a reward of $10$ if it reaches grid 1 or 6, and receives a negative reward of $-1$ if it collides with the other agent. The discount factor being $\gamma=1$. 
Each agent tries to learn an optimal policy based on the observed rewards.
However, the negative reward cannot completely prevent collisions during the learning process of traditional MARL algorithms. 
Because the agents need to explore different (even unsafe) actions to learn about states and rewards from the environment.
Now we show how to construct a shield that can block unsafe actions and guarantee collision free.
We use an observation set $L$ that measures the distance $d$ between blue and orange agents. 
For example, $d=-1$ for agents' positions shown in Figure~\ref{fig:grid}.
We build a coarse environment abstraction DFA $\mathcal{A}^e$ that captures the relation of agents' distances and joint actions.
Figure~\ref{fig:eg_centralized}(a) shows a fragment of $\mathcal{A}^e$.
We can express the safety specification of collision avoidance using the following LTL formula: \\
$\Box \neg \Big((d=0) \vee \big((d=-1) \wedge ((stay, left) \vee (right, left) \vee (right, stay))\big)$ \\
$\vee \big((d=1) \wedge ((stay, right) \vee (left, right) \vee (left, stay))\big)\Big)$ \\
which indicates that the following bad scenarios should never occur:
two agents being in the same grid ($d=0$), or taking certain unsafe joint actions that would make them collide into each other when $d=-1$ or $d=1$.
We can translate the LTL formula into the DFA $\mathcal{A}^s$ shown in Figure~\ref{fig:eg_centralized}(b).
We build a two-player safety game from the product of $\mathcal{A}^e$ and $\mathcal{A}^s$.  Figure~\ref{fig:eg_game} shows a fragment of the safety game. 
For example, in the game state $(q^e_0,q^s_0)$, the blue and orange agents should not choose a joint action (\emph{stay}, \emph{left}) that leads to an unsafe game state $(q^e_1,q^s_1)$ where two agents collide into each other.
The synthesized centralized shield prevents the collision by correcting the unsafe action (\emph{stay}, \emph{left}) with a safe action (\emph{stay}, \emph{stay}) and assigns a punishment cost of $-1$ to the orange agent. 

\startpara{Correctness} 
We show that the synthesized centralized shields can indeed enforce safety specifications for MARL agents as follows. 
Given a trace $s_0a_0s_1a_1 \dots \in (S \times A)^\omega$ jointly produced by MARL agents, the centralized shield, and the environment, there is a corresponding run $q_0q_1\dots \in Q^\omega$ of the shield $\mathcal{S}=(Q, q_0, \Sigma_I, \Sigma_O, \delta, \lambda)$ such that
$q_{i+1} = \delta(q_i, (f(s_i),a_i))$ and $a_i = \lambda(q_i, (f(s_i),a_i))$
for all $i \ge 0$, where $f: S \to L$ is the observation function.
By the construction of the shield, we have $Q = Q^e \times Q^s$, where $Q^e$ and $Q^s$ are the state space of the environment abstraction DFA $\mathcal{A}^e$ and the safety specification DFA $\mathcal{A}^s$, respectively. 
Thus, we can project the run $q_0q_1\dots$ of the shield onto a trace $q^s_0(f(s_0),a_0)q^s_1(f(s_1),a_1)\dots$ on $\mathcal{A}^s$.
The shield is constructed from the winning region of the safety game, which ensures that
only safe states are ever visited along the trace $q^s_0(f(s_0),a_0)q^s_1(f(s_1),a_1)\dots$ of $\mathcal{A}^s$ (i.e., $q^s_i \in F^s$ for all $i \ge 0$).
Thus, the centralized shield $\mathcal{S}$ can guarantee that the safety specification $\mathcal{A}^s$ is never violated.

\startpara{Impact on Learning Performance}
The centralized shielding approach is agnostic to the choice of a MARL algorithm, because the shield interacts with the learner only via inputs and outputs, and does not rely on the inner-workings of the learning algorithm.
As explained in Section~\ref{sec:background}, there is a lack of theoretical convergence guarantees for MARL algorithms in general. 
Thus, a full theoretical analysis of the shielding approach's impact on MARL convergence is out of scope for this paper. 
We show empirically in our experiments (Section~\ref{sec:exp}) that 
(1) MARL with and without centralized shielding both converge;
(2) centralized shielding can guarantee the safety in all examples, while MARL without shielding does not prevent agents' unsafe behavior;
(3) centralized shielding learns more optimal policies with better returns than non-shielded MARL in some examples (e.g., due to the removal of unsafe actions that may destabilize learning).

	\section{Factored Shielding}
	
\begin{figure}[t]
	\centering
	\includegraphics[width=\columnwidth]{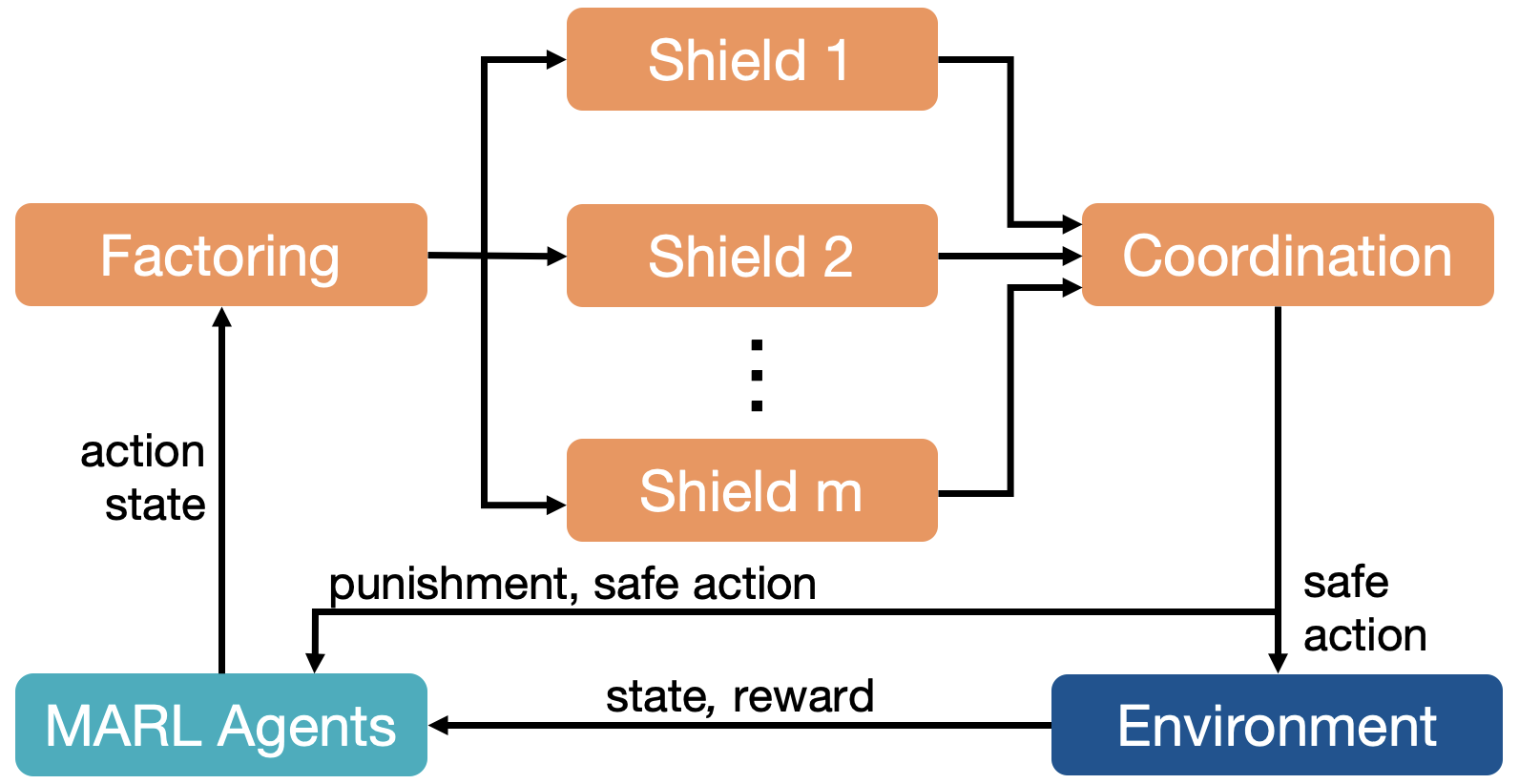}
	\caption{Safe MARL with factored shielding.}
	\label{fig:factored}
\end{figure}

\noindent
The centralized shielding approach has limited scalability, because the computational cost of shield synthesis grows exponentially with the number of agents. 
To address this limitation, we develop a \emph{factored shielding} approach that synthesizes multiple shields to monitor MARL agents concurrently,
as illustrated in Figure~\ref{fig:factored}.

Let us consider a finite set of factored shields $\{\mathcal{S}_1, \cdots, \mathcal{S}_m\}$ where each shield is synthesized based on a factorization of the joint state space observed by all agents.  
We can leverage problem-specific knowledge to achieve an efficient factorization scheme (e.g., how many shields to use, what is the state space covered by each shield).
For example, we synthesize two factored shields $\mathcal{S}_1$ and $\mathcal{S}_2$ for monitoring agents' behavior in grids 1-3 and 4-6 of Figure~\ref{fig:grid}, respectively. 
A factored shield monitors a subset of agent actions at each time step. 
A shield is not tied to any specific agent; instead, an agent can request to join or leave a shield from border states at any time. 
For example, if the orange agent in Figure~\ref{fig:grid} wants to move from grid 4 to grid 3, it would request to join $\mathcal{S}_1$ and leave $\mathcal{S}_2$.



Algorithm~\ref{code_factored} describes how the factored shielding works at each time step $t$. 
There are three phases: (1) factorization, (2) shielding, and (3) coordination. 
In the factorization phase (line 5-14), the algorithm identifies the factored shields that are responsible for monitoring each agent $k$ in the current time step $t$, based on a mapping between the agent state $s^k_t$ and the factored state space assigned to each shield $\mathcal{S}_i$.
Thus, there must exist at least one factored shield monitoring each agent. 
If agent $k$ happens to be in a border state $s^k_t$ within the shield $\mathcal{S}_i$ and, by taking action $a^k_t$, the agent would cross the border to another shield $\mathcal{S}_j$, the algorithm relates agent $k$ with both shields and renames its actions in shield $\mathcal{S}_i$ and $\mathcal{S}_j$ as \emph{leave} and \emph{join}, respectively.   
Next, in the shielding phase (line 16-33), each factored shield checks if the set of related agents act safely (i.e., not violating the safety specification within it) and substitutes any unsafe action with a default safe action (e.g., \emph{stay} in our running example).
In the coordination phase (line 35-47), the algorithm checks the output of all shields to make sure compatible decisions are made for each agent. 
For example, if an agent action $a^k_t$ is translated to requests of leaving $\mathcal{S}_i$ and joining $\mathcal{S}_j$, then both requests need to be approved by the shields; however, if $\mathcal{S}_j$ considers \emph{join} as unsafe at this time and substitutes with a default safe action \emph{stay}, then the algorithm corrects the agent action $a^k_t$ and output with safe action $\bar{a}^k_t=stay$,
Finally, the algorithm assigns a punishment cost $\rho^k_t = c$ for any unsafe action $a^k_t$ with 
$\bar{a}^k_t \neq a^k_t$.

\begin{algorithm}[t]
\caption{Factored shielding at time step $t$} \label{code_factored}
\algsetup{linenosize=\tiny}
\small
\begin{algorithmic} [1]
\REQUIRE A set of factored shields $\{\mathcal{S}_1,\cdots,\mathcal{S}_m\}$, 
MARL agents' joint action $a_t=(a^1_t,\cdots,a^n_t)$ and joint state $s_t=(s^1_t,\cdots,s^n_t)$,
a default safe action $b$, a constant punishment cost $c$
\ENSURE Safe joint action $\bar{a}_t$, punishment $\rho_t$
\STATE {Initialize int array $\mathsf{A2S}: n \times 2 $ \textcolor{teal}{// related shield index} }
\STATE {Initialize string array $\mathsf{Act}: n \times 2 $ \textcolor{teal}{// actions} }
\STATE {Initialize Boolean array $\mathsf{S2A}: m \times n $ \textcolor{teal}{// agents in each shield} }
\STATE {\textcolor{teal}{ // Factorization phase}}
\FORALL{agent $k \in \{1,\cdots,n\}$ } 
    \STATE find a factored shield $\mathcal{S}_i$ related to the agent state $s^k_t$
    \IF{ $(s^k_t,a^k_t)$ may leave shield $\mathcal{S}_i$ and join shield $\mathcal{S}_j$}
        \STATE $\mathsf{A2S}[k][0] \leftarrow i$, $\mathsf{A2S}[k][1] \leftarrow j$
        \STATE $\mathsf{Act}[k][0] \leftarrow ``leave"$, $\mathsf{Act}[k][1] \leftarrow ``join"$
        \STATE $\mathsf{S2A}[i][k] \leftarrow \mathsf{True}$, $\mathsf{S2A}[j][k] \leftarrow \mathsf{True}$
    \ELSE
        \STATE $\mathsf{A2S}[k][0] \leftarrow i$,
            $\mathsf{Act}[k][0] \leftarrow a^k_t$,
            $\mathsf{S2A}[i][k] \leftarrow \mathsf{True}$
    \ENDIF
\ENDFOR 
\STATE {\textcolor{teal}{ // Shielding phase}}
\FORALL{shield $\mathcal{S}_i$ with $i \in \{1,\cdots,m\}$ }
    \STATE $a \leftarrow \{\} $ 
    \FORALL{$k$ with $\mathsf{S2A}[i][k] = \mathsf{True}$}
        \IF{$\mathsf{A2S}[k][0] = i$}
            \STATE $a \leftarrow $ append $\mathsf{Act}[k][0]$
        \ELSE
            \STATE $a \leftarrow $ append $\mathsf{Act}[k][1]$
        \ENDIF
    \ENDFOR
        \STATE $\bar{a} \leftarrow$ safe action output by the shield $\mathcal{S}_i$
        \FORALL{agent $k$ such that $\bar{a}^k \neq a^k$}
            \IF{$\mathsf{A2S}[k][0] = i$}
            \STATE $\mathsf{Act}[k][0] \leftarrow \bar{a}^k$
        \ELSE
            \STATE $\mathsf{Act}[k][1] \leftarrow \bar{a}^k$
        \ENDIF
        \ENDFOR
\ENDFOR
\STATE {\textcolor{teal}{ // Coordination}}
\FORALL{agent $k \in \{1,\cdots,n\}$ } 
    \IF{$\mathsf{A2S}[k][1] \neq null$}
        \IF{$\mathsf{Act}[k][0] = ``leave"$ and $\mathsf{Act}[k][1] = ``join"$}
            \STATE $\mathsf{Act}[k][0] \leftarrow a^k_t$
        \ELSE
            \STATE $\mathsf{Act}[k][0] \leftarrow b$ 
        \ENDIF
    \ENDIF
    \STATE $\bar{a}^k_t \leftarrow \mathsf{Act}[k][0]$, $\rho^k_t \leftarrow 0$
    \IF{$\bar{a}^k_t \neq a^k_t$}
        \STATE $\rho^k_t \leftarrow c$ 
    \ENDIF
\ENDFOR 
\RETURN $\bar{a}_t$, $\rho_t$
\end{algorithmic}
\end{algorithm}

We synthesize factored shields using a similar method as the synthesis of centralized shields.
However, instead of building a safety game that accounts for the joint states $S$ and joint actions $A=A^1 \times \cdots \times A^n$ of all MARL agents, we only consider a factorization of states and actions for the synthesis of each factored shield.
Let $S_i \subseteq S$ be the factored state space of shield $\mathcal{S}_i$. 
We factor the coarse environment abstraction DFA $\mathcal{A}^e$ into a DFA $\mathcal{A}^e_i=(Q^e_i, q^e_{0,i}, \Sigma^e_i, \delta^e_i, F^e_i)$ with the alphabet $\Sigma^e_i = L_i \times A_i$, where an observation function $f: S_i \to L_i$ maps the factored states $S_i$ to some observation set $L_i$, and $A_i={(A^1 \cup \cdots \cup A^n \cup \{join, leave\})} \times \dots \times {(A^1 \cup \cdots \cup A^n \cup \{join, leave\})}$ is the joint action in shield $\mathcal{S}_i$ with $|A_i|$ determined by the maximum number of agents that shield $\mathcal{S}_i$ can monitor at once. 
Note that we need to translate the agent actions at border states of a shield to \emph{join} or \emph{leave} requests. 
Intuitively, since any agent may request to join or leave shield $\mathcal{S}_i$ at any time, the joint action $A_i$ needs to account for any possible combination of agents.
This allows us to synthesize factored shields offline with a fixed alphabet, instead of re-computing shields for different agents at each step during learning. 
Similarly, we can factor the safety specification DFA $\mathcal{A}^s=(Q^s, q^s_0, \Sigma^s, \delta^s, F^s)$ into a DFA $\mathcal{A}^s_i=(Q^s_i, q^s_{0,i}, \Sigma^s_i, \delta^s_i, F^s_i)$ with the alphabet $\Sigma^s_i = L_i \times A_i$.
We obtain the shield $\mathcal{S}_i$ as a Mealy machine by solving the two-player safety game $\mathcal{G}_i$ built from $\mathcal{A}^e_i$ and $\mathcal{A}^s_i$, in a similar way as described in Section~\ref{sec:centralized}.

\begin{figure}[t]
	\centering
	\includegraphics[width=0.6\columnwidth]{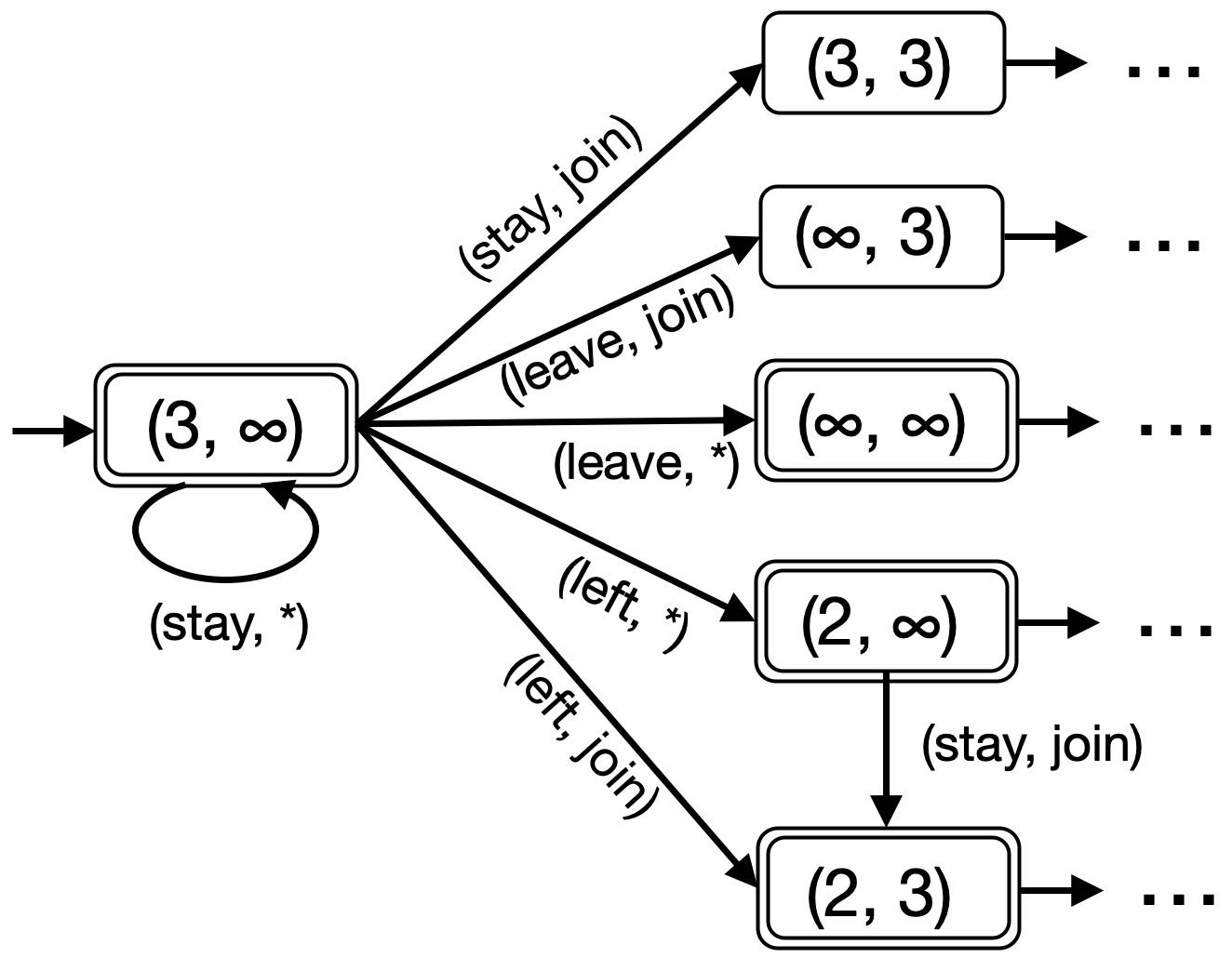}
	\caption{An excerpt of the safety game for constructing shield $\mathcal{S}_1$ of our running example. Double lines indicate safe states. To simplify the graphic notation, we put observations inside each state which should be labeled on all outgoing transitions from that state. The observations are about agents' grid positions, with $\infty$ denoting outside. $*$ refers to any action except ``join''. }
	\label{fig:eg_factored}
\end{figure}

Figure~\ref{fig:eg_factored} shows an example safety game for synthesizing the shield $\mathcal{S}_1$ that monitors agents' actions in grid 1-3 of our running example. 
The initial game state observes that the blue agent is in grid 3 and the orange agent is outside the shield. 
If the blue and orange agents ask for a pair of actions (\emph{stay}, \emph{join}), then the game would move to an unsafe state where both agents collide into each other in grid 3.
In this case, shield $\mathcal{S}_1$ substitutes (\emph{stay}, \emph{join}) with safe actions (\emph{stay}, \emph{stay}).
Since the orange agent is involved in two shields $\mathcal{S}_1$ and $\mathcal{S}_2$, we need to coordinate the output of both shields.
For example, if $\mathcal{S}_1$ rejects orange agent's \emph{join} request but $\mathcal{S}_2$ accepts the same agent's \emph{leave} request, then there is conflict among the output of $\mathcal{S}_1$ and $\mathcal{S}_2$.
In such case, our coordination algorithm chooses the default safe action \emph{stay} for the orange agent.
Note that, if there is another agent in shield $\mathcal{S}_2$, then it should not be allowed to move to grid 4 before the orange agent successfully leaves $\mathcal{S}_2$ to avoid collision. 
Such safety constraints can be encoded in the safety game for synthesizing the shield $\mathcal{S}_2$.

\startpara{Correctness} 
We show that the factored shielding algorithm can guarantee safety for MARL agents. 
Given a trace $s_0a_0s_1a_1 \dots \in (S \times A)^\omega$ jointly produced by MARL agents, the factored shielding, and the environment, we prove that the state-action pair $(s_t, a_t)$ is safe at every time step $t$. There are several cases.
First, suppose none of the agents requests to switch shields at time step $t$.
By the construction of factored shields, each shield $\mathcal{S}_i$ monitors a subset of agents based on the factored state space $s_{t,i}$ and outputs a safe joint action $a_{t,i}$ that does not violate the safety specification. 
Thus, the joint state $s_t=s_{t,1}\cup \cdots \cup s_{t,m}$ and joint action $a_t=a_{t,1} \cup \cdots \cup a_{t,m}$ output by all shields are safe for all agents. 
Second, suppose there is some agent $k$ requesting to leave shield $\mathcal{S}_i$ and join shield $\mathcal{S}_j$.
If both shields accept agent $k$'s requests, which means that agent $k$ does not cause a violation of safety specification with either shield. So we still have $s_t$ and $a_t$ safe for all agents. 
If $\mathcal{S}_j$ rejects agent $k$'s joining request and substitutes with a default safe action, 
then the factored shielding algorithm coordinates with the output of $\mathcal{S}_i$ and corrects agent $k$'s leaving request with the default safe action as well. 
Such a correction does not affect the safety of other agents in shield $\mathcal{S}_i$, because by construction the shield accounts for the worst case scenario of leaving request being rejected. 
Therefore, we have the joint state-action pair $(s_t, a_t)$ safe at every time step $t$ for all agents.

\startpara{Impact on Learning Performance} 
Similarly to centralized shielding, the factored shielding approach is agnostic to the choice of a MARL algorithm.
We show empirically via our experiments that adding factored shields does not prevent MARL algorithms from converging. In addition, our experiments show that the factored shielding approach can be applied to examples where the synthesis of centralized shields is not feasible due to a large number of agents.
While the two shielding approaches can both guarantee the safety during learning in all examples, factored shielding sometimes leads to less optimal policies than centralized shielding (e.g., due to the delay caused by agents switching shields).

	\section{Experiments} \label{sec:exp}
	
\begin{figure}[th]
\centering
\includegraphics[width=65mm,,, scale=1]{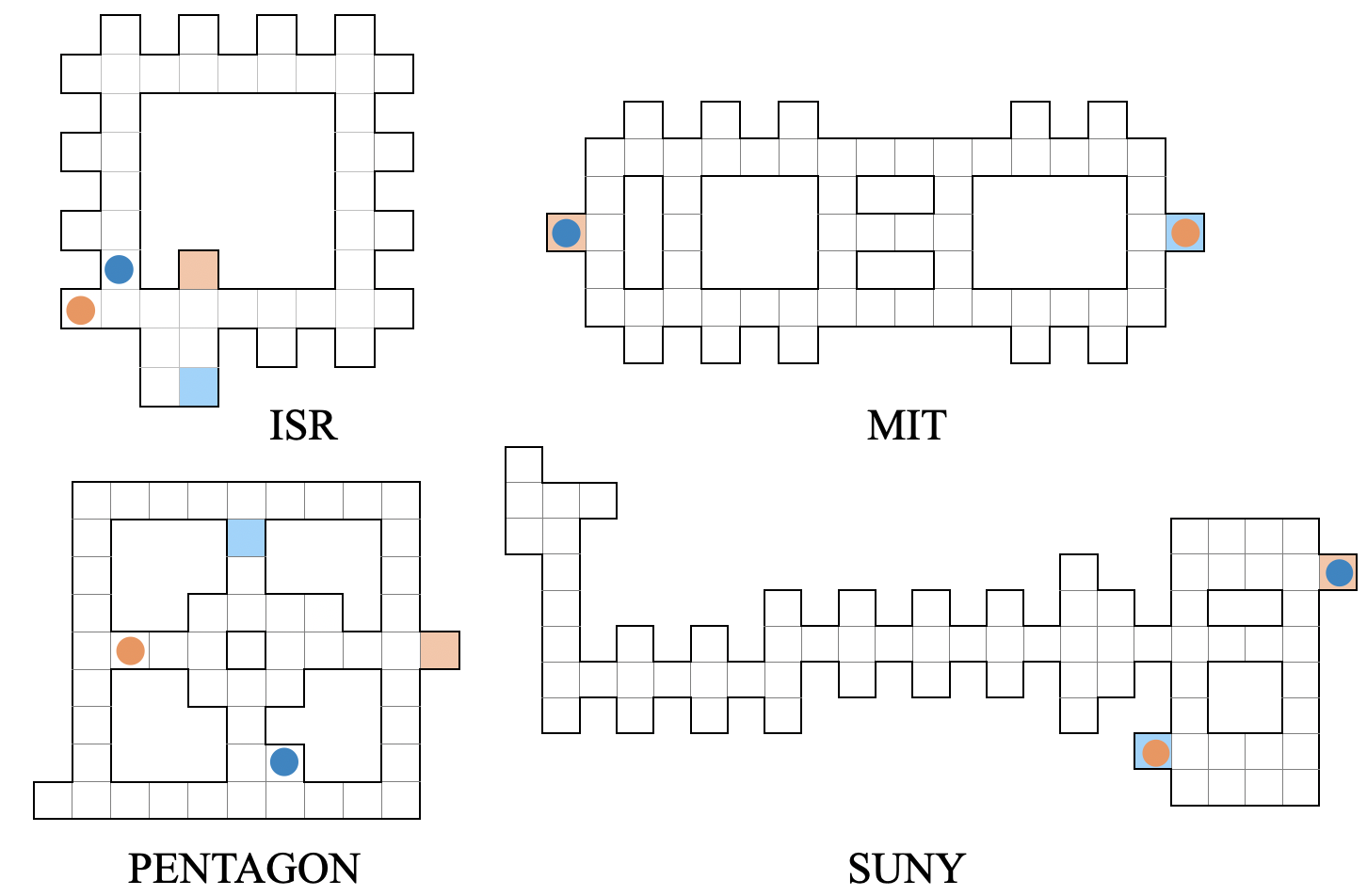}
\caption{Maps of grid world examples adapted from~\cite{melo2009learning}. In each map, blue and orange agents aim to learn optimal policies to navigate from start (circles) to target (squares) while avoiding collisions.}
\label{fig:maps}
\end{figure}

\begin{figure}[th]
    \centering
    \includegraphics[,,, scale=0.20]{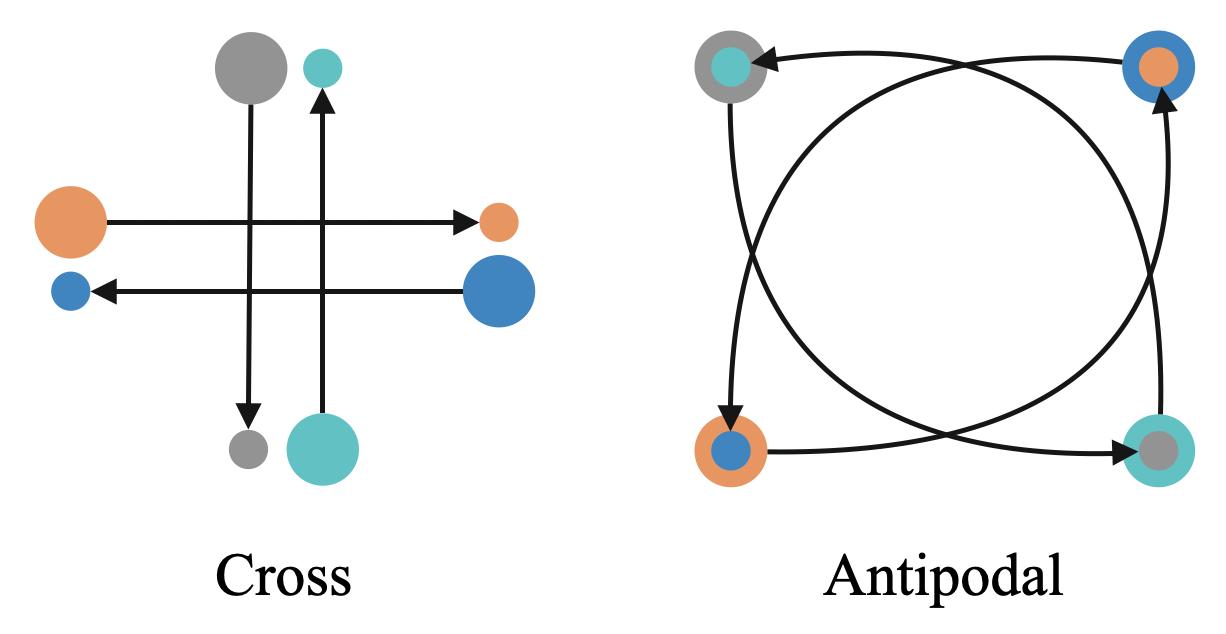}
    \caption{Visualisations of cooperative navigation examples adapted from~\cite{yang2019cm3}. Four agents (blue, orange, green, and grey) aim to learn optimal policies to navigate from start (large circles) to target (small circles) while avoiding collisions.}
    \label{fig:particle_env}
\end{figure}

We implemented both the centralized shielding and factored shielding approaches in Python and used the Slugs tool \cite{ehlers2016slugs} to synthesize shields via solving two-player safety games.
We applied our prototype implementation to six benchmark problems in the grid world (Figure~\ref{fig:maps}) and a cooperative navigation environment (Figure~\ref{fig:particle_env}).
We used two MARL algorithms CQ-learning~\cite{de2010learning} and MADDPG~\cite{lowe2017multi} in experiments to show that our shielding approaches are agnostic to the choice of MARL algorithms. 
The experiments were run on a computer with Intel i5 CPU and 16 GB of RAM. 
Each experiment was split into training phase (linearly decreasing exploration) and evaluation phase (immediately following the training phase and with an exploration rate of 5\%). 
All experiments were conducted for 10 independent runs whose results were averaged to reduce the impact of outliers.
The shields in all examples were synthesized within two minutes.

\startpara{Problem Setup}
Figure~\ref{fig:maps} shows four maps of benchmark grid world examples adapted from~\cite{melo2009learning}. Each map has two agents, where each agent aims to learn its own optimal policy for navigating from the start position to the target position while trying to avoid collisions. Each agent has five possible actions: \emph{stay}, \emph{up}, \emph{down}, \emph{left}, \emph{right}. Once an agent reaches its target position, it stays there. 
A learning episode ends when both agents have reached their target positions. 
Both agents have the same reward function: $-1$ for a valid move, $-10$ for a collision with a wall, $-30$ for collision with the other agent, $100$ for arriving at the agent's target position. 

Figure~\ref{fig:particle_env} shows two benchmark cooperative navigation examples adapted from~\cite{yang2019cm3}. Each example has four agents represented as particles. The goal is for agents to cooperate and reach their designated target positions as fast as possible while avoiding collisions. 
We discretize the fully continuous environment in \cite{yang2019cm3} by restricting agents only take positions with a precision of $0.1$.
An agent receives a higher reward when it gets closer to its target position (i.e., negation of the distance value), and a negative reward $-1$ for any collision.

 \begin{figure}[t]
    \centering
    \includegraphics[,,, scale=0.38]{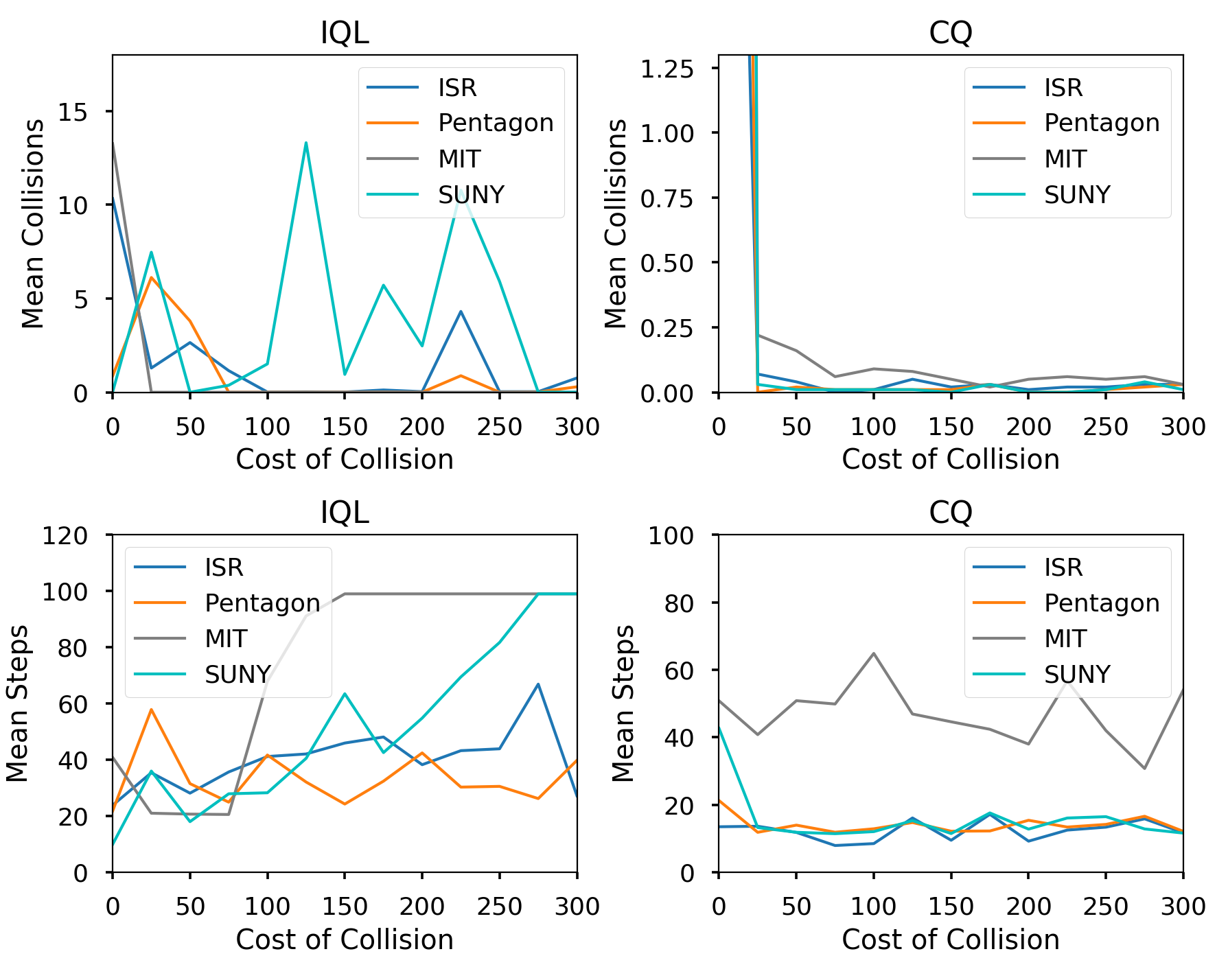}
    \caption{Collision variation experiments results (average of 10 evaluation episodes conducted after 1,000 training episodes, across 10 independent runs).}
    \label{fig:collision_var}
\end{figure}

\startpara{Collision Variation Experiments}
We conducted a set of experiments using the grid world examples to highlight why relying on the reward function only is not sufficient to achieve safety (i.e., collision avoidance in our examples).
To prevent collisions, the traditional practice of reinforcement learning is to assign a negative reward (we refer to its absolute value as the cost of collision) whenever a collision occurs, and increase the cost until the probability of collision happening becomes negligible. 
Figure~\ref{fig:collision_var} shows the results of our experiments using the independent Q-learning~\cite{tan1993multi} and CQ-learning\cite{de2010learning}.
The left side of the figure shows that, for the independent Q-learning, increasing the cost of collision cannot guarantee that the evaluation phase will be completely collision free; moreover, the increased cost of collision leads to a significant agent performance degradation measured by a larger number of steps to reach target positions. 
In the MIT and SUNY maps, agents even learn policies that give up the primary task of reaching target positions in order to avoid the high collision cost. 
The results of the CQ-learning (shown in the right side of the figure) are better than those of the independent Q-learning.
The number of collisions drops quickly with a relatively low cost. However, CQ-learning cannot guarantee zero collision either (see Table~\ref{results_combined}).

\begin{table*}[ht]
\setlength{\tabcolsep}{0.25em}
\centering
\begin{tabular}{|lc|ccc|ccc|ccc|ccc|}
\hline
\multicolumn{1}{|c}{}          & \multicolumn{1}{l|}{}        & \multicolumn{3}{c|}{IQL}       & \multicolumn{3}{c|}{CQ}     & \multicolumn{3}{c|}{CQ with centralized shield} & \multicolumn{3}{l|}{CQ with factored shield} \\ \hline
\multicolumn{1}{|c|}{Maps}     & \multicolumn{1}{l|}{Optimal Steps} & Steps & Reward   & Collisions & Steps & Reward & Collisions & Steps        & Reward        & Collisions       & Steps       & Reward       & Collisions      \\ \hline
\multicolumn{1}{|l|}{ISR}      & 5                            & 30.35 & -10.20   & 20.30      & 8.66  & 89.53  & 0.40       & {7.03}         & {93.85}         & \textbf{0.00}             & {7.31}        & {93.74}        & \textbf{0.00}            \\
\multicolumn{1}{|l|}{Pentagon} & 10                           & 46.58 & -19.17   & 11.60      & {10.96} & {88.96}  & 0.20       & 12.08        & 88.44         & \textbf{0.00}             & 13.20       & 84.88        & \textbf{0.00}            \\
\multicolumn{1}{|l|}{MIT}      & 18                           & {20.84} & {77.33}    & {0.00}       & 42.93 & 30.38  & 0.90       & {28.38}        & {73.94}         & \textbf{0.00}             & 29.96       & 37.96        & \textbf{0.00}            \\
\multicolumn{1}{|l|}{SUNY}     & 10                           & 34.80 & -160.175 & 72.60      & 13.97 & 84.78  & 0.30       & {11.97 }       & {88.44}         & \textbf{0.00}             & 14.02       & 83.77        & \textbf{0.00}            \\ \hline
\end{tabular}
\caption{Results comparing the independent Q-learning, CQ-learning, CQ-learning with centralized and factored shields (average of 10 evaluation episodes conducted after 1,000 training episodes, across 10 independent runs).}
\label{results_combined}
\end{table*}

\begin{figure}[b]
    \centering
    \includegraphics[,,, scale=0.39]{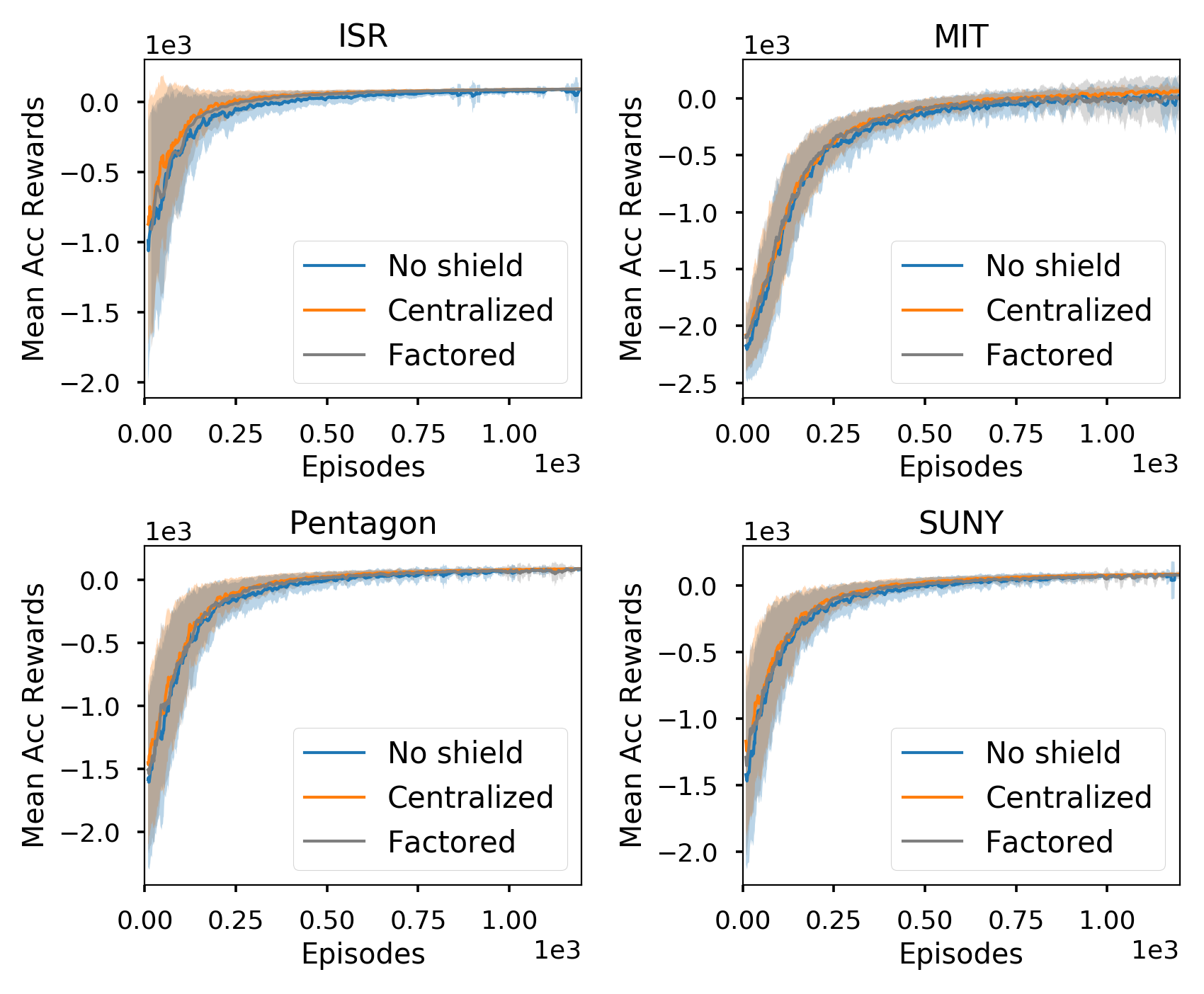}
    \caption{Comparison of CQ-learning without shielding, with centralized or factored shielding based on the accumulated rewards per episode (average and standard deviation over 1,000 training episodes, across 10 independent runs).}
    \label{fig:res_curves_gridworld}
\end{figure}

\startpara{Centralized Shielding Evaluation}
We integrated CQ-learning with centralized shielding and applied it to the four grid world examples shown in Figure~\ref{fig:maps}.
The results in Table~\ref{results_combined} show that centralized shielding can guarantee collision free learning in all cases. 
Moreover, in three out of four maps, CQ-learning with centralized shield obtained better policies with higher rewards and smaller number of steps to reach the target, compared to no shielding. 
Figure~\ref{fig:res_curves_gridworld} shows that centralized shielding achieves the highest accumulated reward in most times; moreover, the blue shaded area (standard deviation of no shielding) tends to stretch lower than others, indicating that CQ-learning without shielding obtains lower rewards than with centralized shielding on average.
The learning curves also show that the centralized shielding does not prevent the learner from converging across different examples. 
However, we failed to synthesize centralized shields with more than two agents in these grid maps, due to scalability issues of shield synthesis.

 \begin{figure}[b]
    \centering
    \includegraphics[,,, scale=0.43]{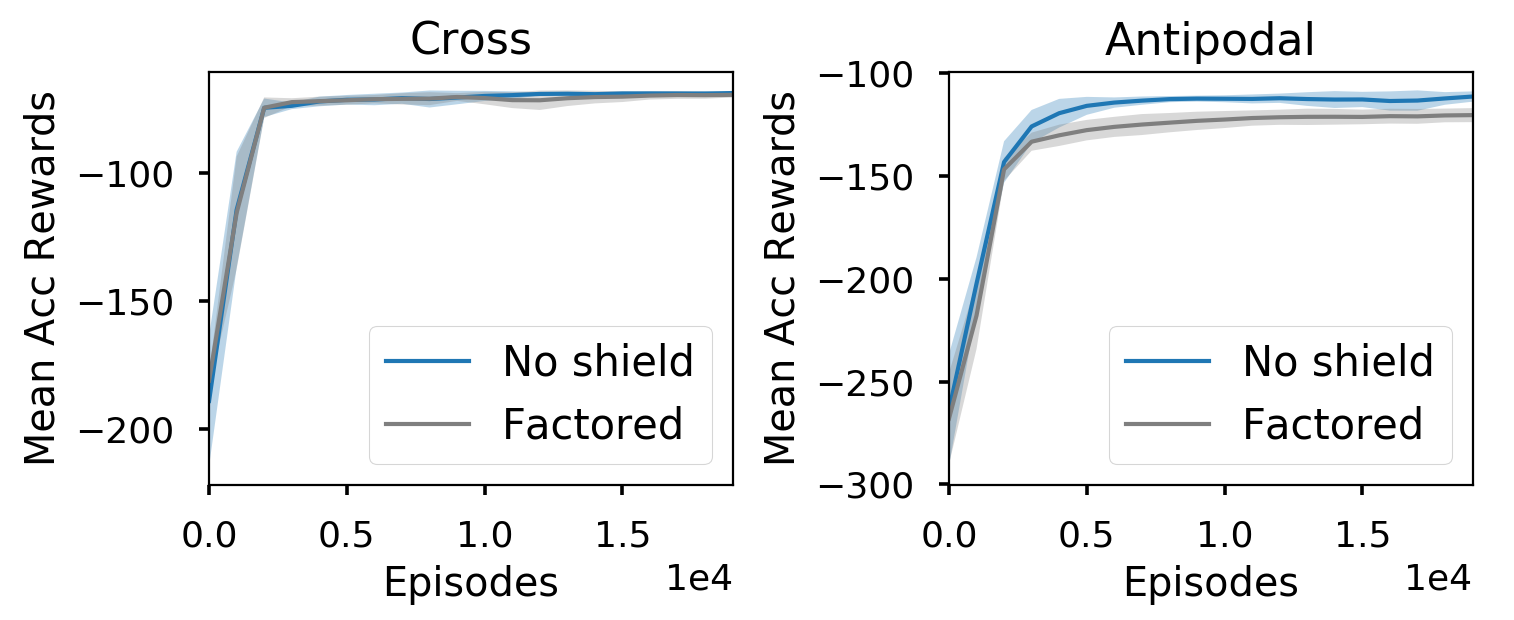}
    \caption{Comparison of MADDPG without and with factored shielding based on the accumulated rewards per episode (average and standard deviation over 20,000 training episodes, across 10 independent runs).}
    \label{fig:deep_rl}
\end{figure}

\begin{table}[b]
\centering
\begin{tabular}{|l|c|c|}
\hline
          & MADDPG    & MADDPG with Shield \\ \hline
Cross     & 207.20    & \textbf{0.00}               \\ \hline
Antipodal & 14,419.20 & \textbf{0.00}               \\ \hline
\end{tabular}
\caption{Total number of collisions over 20,000 training episodes for the cooperative navigation examples.}
\label{tab:dmarl_collisions}
\end{table}

\startpara{Factored Shielding Evaluation}
First, we applied CQ-learning with factored shielding to the four grid world examples. We adopted a factorization scheme such that each shield monitors agent actions occurring within a $3\times3$ grid block in each map. 
Results in Table~\ref{results_combined} show that CQ-learning with factored shielding can guarantee zero collisions in all examples, while learned policies have similar quality as those obtained from CQ-learning with centralized shielding.
Figure~\ref{fig:res_curves_gridworld} shows that factored shielding achieves similar performance in terms of the accumulated rewards per episode, compared to centralized shielding and without shielding.
Due to the scalability limitation of CQ-learning, we can only consider two agents in these examples. 

Additionally, we integrated a different algorithm MADDPG~\cite{lowe2017multi} with factored shielding and applied it to the cooperative navigation examples shown in Figure~\ref{fig:particle_env} with a $5 \times 5$ shield size where one unit of distance corresponds to $0.1$ in the environment. There are four agents in each example, which is not feasible for centralized shielding approach to handle.
Table~\ref{tab:dmarl_collisions} shows that MADDPG with factored shielding can guarantee zero collisions over the training period of $20,000$ episodes for both examples.
By contrast, MADDPG without shielding leads to about $207$ and $14,419$ occurrences of collisions for the cross and antipodal examples, respectively. 
Figure~\ref{fig:deep_rl} shows that in the cross example, MADDPG without and with factored shielding have comparable learning performance in terms of the accumulated rewards per episode; in the antipodal example, MADDPG without shielding achieves higher rewards than MADDPG with factored shielding, though this comes at a trade-off of more collisions. 
The learning curves in Figure~\ref{fig:deep_rl} also show that the factored shielding do not have negative impact on the learner's ability to converge.

\startpara{Summary}
Our experiments demonstrate that the two shielding approaches can guarantee the safety, without compromising the learning performance in terms of the convergence rate and the quality of learned policies. 
Moreover, factored shielding is more scalable in the number of agents than centralized shielding.\\
	
	\section{Conclusion}
	In this paper, we present two shielding approaches that guarantee the safety specifications expressed in linear temporal logic (LTL) during the learning process of MARL. 
The centralized shielding approach synthesizes a single shield to centrally monitor the joint actions of all agents and only corrects any unsafe action that violates the LTL safety specification. 
However, the scalability of centralized shielding is restricted because the computational cost of shield synthesis grows exponentially with the number of agents. 
The factored shielding approach addresses this limitation by synthesizing multiple factored shields with each shield monitoring a subset of agents at each time step. 
Our experimental results show that both shielding approaches can guarantee the safety specification (e.g., collision avoidance) during learning, and achieve similar learning performance (e.g., convergence speed, quality of learned policies) as non-shielded MARL. 
We manually devise factorization schemes for the factored shielding approach in our experiments based on the problem-specific knowledge. 
In the future, we will explore the automated learning of efficient factorization schemes.


	\section{Acknowledgements}
    This work was supported in part by ONR grant N00014-18-1-2829 and ARO grant W911NF-20-1-0140. 
    Any opinions, findings, and conclusions or recommendations expressed in this material are those of the author(s) and do not necessarily reflect the views of the grant sponsors.

	\medskip
	\bibliography{references}  
	\bibliographystyle{ACM-Reference-Format}
	
\end{document}